\documentclass[conference]{IEEEtran}
\pdfoutput=1
\usepackage{graphicx}
\usepackage{shortcuts}
\usepackage{xspace}
\usepackage{code}
\usepackage{cite}
\usepackage{subcaption}
\usepackage{footnote}
\usepackage{mathtools}
\usepackage{chngcntr}
\usepackage{listings, color, caption}
\usepackage{threeparttable}
\usepackage{setspace}
\usepackage{multirow}
\usepackage{multicol}
\usepackage{pifont}
\usepackage{soul}
\usepackage{boxedminipage}
\usepackage{array}
\usepackage{siunitx} % for decimal alignment in tables
\usepackage{algorithm,algorithmicx}
\usepackage[noend]{algpseudocode}
\usepackage{comment}
\graphicspath{ {images/} }
\usepackage{amsfonts}
\usepackage{mathrsfs}
\usepackage{amssymb}
\usepackage{amsthm}
\usepackage{hyperref}
\definecolor{gray}{rgb}{0.4,0.4,0.4}
\definecolor{darkblue}{rgb}{0.0,0.0,0.6}
\definecolor{cyan}{rgb}{0.0,0.6,0.6}
\usepackage{url}

\DeclareRobustCommand*{\IEEEauthorrefmark}[1]{%
    \raisebox{0pt}[0pt][0pt]{\textsuperscript{\footnotesize\ensuremath{#1}}}}

\lstdefinelanguage{XML}
{
  morestring=[b]",
  morestring=[s]{>}{<},
  morecomment=[s]{<?}{?>},
  stringstyle=\color{black},
  identifierstyle=\color{darkblue},
  keywordstyle=\color{cyan},
  morekeywords={xmlns,version,type}% list your attributes here
}
\lstdefinelanguage{JavaScript}{
    keywords={typeof, new, true, false, catch, function, return, null, catch, switch, var, if, in, while, do, else, case, break},
    keywordstyle=\color{blue}\bfseries,
    ndkeywords={class, export, boolean, throw, implements, import, this},
    ndkeywordstyle=\color{green}\bfseries,
    identifierstyle=\color{black},
    sensitive=false,
    comment=[l]{//},
    morecomment=[s]{/*}{*/},
    commentstyle=\color{purple}\ttfamily,
    stringstyle=\color{blue}\ttfamily,
    morestring=[s]{`}{'},
}
\definecolor{lightgray}{rgb}{.9,.9,.9}
\definecolor{darkgray}{rgb}{.4,.4,.4}
\definecolor{purple}{rgb}{0.65, 0.12, 0.82}
\newcommand{\RN}[1]{%
  \textup{\uppercase\expandafter{\romannumeral#1}}%
}

\algrenewcommand\algorithmicindent{0.7em}%

\title{VMID: A Multimodal Fusion LLM Framework for Detecting and Identifying Misinformation of Short Videos}
\author{
    \IEEEauthorblockN{Weihao Zhong\IEEEauthorrefmark{1}, Yinhao Xiao\IEEEauthorrefmark{2}\IEEEauthorrefmark{*},
        Minghui~Xu\IEEEauthorrefmark{3}\IEEEauthorrefmark{*}, Xiuzhen Cheng\IEEEauthorrefmark{3},\textit{~Fellow,~IEEE}}

    \IEEEauthorblockA{\IEEEauthorrefmark{1,2}School of Information Science, Guangdong University of Finance and Economics, 
    \\Guangdong Intelligent Business Engineering Technology Center, 
    \\Key Laboratory of Collaborative Innovation in Digital Economy,Guangzhou, China.
    \\ zhongweihao@student.gdufe.edu.cn, 20191081@gdufe.edu.cn}
    
    \IEEEauthorblockA{\IEEEauthorrefmark{3}School of Computer Science, Shandong University, Qingdao, China.
    \\ \{mhxu,xzcheng\}sdu.edu.cn}
    
    \IEEEauthorblockA{Corresponding Author: Yinhao Xiao and Minghui Xu}

}

\begin{document}

\maketitle
% \begin{abstract}
\section{abstract}
Short video platforms have become important channels for news dissemination, offering a highly engaging and immediate way for users to access current events and share information. However, these platforms have also emerged as significant conduits for the rapid spread of misinformation, as fake news and rumors can leverage the visual appeal and wide reach of short videos to circulate extensively among audiences. Existing fake news detection methods mainly rely on single-modal information, such as text or images, or apply only basic fusion techniques, limiting their ability to handle the complex, multi-layered information inherent in short videos. To address these limitations, this paper presents a novel fake news detection method based on multimodal information, designed to identify misinformation through a multi-level analysis of video content. This approach effectively utilizes different modal representations to generate a unified textual description, which is then fed into a large language model for comprehensive evaluation. The proposed framework successfully integrates multimodal features within videos, significantly enhancing the accuracy and reliability of fake news detection. Experimental results demonstrate that the proposed approach outperforms existing models in terms of accuracy, robustness, and utilization of multimodal information, achieving an accuracy of 90.93\%, which is significantly higher than the best baseline model (SV-FEND\cite{Qi2023FakeSV}) at 81.05\%. Furthermore, case studies provide additional evidence of the effectiveness of the approach in accurately distinguishing between fake news, debunking content, and real incidents, highlighting its reliability and robustness in real-world applications.
% \end{abstract}

\section{Introduction}

With the rise of social media platforms such as Twitter and Weibo in China, these platforms have become essential channels for people to access the latest news and freely express their opinions \cite{smith_anderson_2018}. However, the convenience and openness of social media have also facilitated the rapid spread of misinformation, which includes news containing intentionally false information. Misinformation not only disrupts the order of cyberspace but also has significant negative impacts on real-world events. For instance, in the political domain, during the month leading up to the 2016 U.S. presidential election, Americans were exposed to an average of one to three pieces of fake news from well-known publishers \cite{silverman_2016}, which inevitably misled voters and influenced the election outcome. In the economic domain, a piece of fake news claiming that Barack Obama had been injured in an explosion caused a loss of \$130 billion in stock value \cite{barron_2013}. In the social domain, in India, numerous innocent people were killed by locals due to a widely circulated fake news story about child trafficking \cite{chauhan_2018}. Similarly, during the early stages of the COVID-19 pandemic in China, rumors spread widely on social media platforms like Weibo, including false claims that drinking strong alcohol could prevent infection or that certain medicines could cure the virus \cite{10.3389/fpubh.2022.864955}. These rumors fueled widespread panic and misinformation, complicating public health responses and delaying effective measures. Therefore, the automatic detection of fake news has become an urgent and critical issue in recent years \cite{8397048}.

The advancement of multimedia technology has propelled the evolution of user-generated news, transforming it from text-based posts to multimedia posts incorporating images and videos. This shift has garnered greater consumer attention and enabled more credible storytelling. On the one hand, visual content, such as images and videos, is more engaging and attention-grabbing than plain text, thus accelerating the spread of news. For example, tweets that include images receive 18\% more clicks, 89\% more likes and 150\% more retweets compared to those without images \cite{invidplugin}. On the other hand, visual content is often used as evidence to support a narrative, enhancing its credibility. Unfortunately, this advantage is also exploited by fake news. In order to spread rapidly, fake news often features misrepresented or even tampered images or videos to attract and mislead consumers. As a result, visual content has become a key element of fake news, making multimedia fake news detection an emerging challenge.

Existing fake news detection methods have primarily focused on text-based or text-image combined news. In contrast, the detection of fake news in videos remains a relatively emerging and under-explored area of research. Traditional video-based fake news detection methods rely on single-modal information, such as extracting text or image features from videos for detection \cite{10.1145/1963405.1963500, 10.1145/3410566.3410599}. These methods often overlook the synergistic effect of multimodal information within the video, failing to effectively combine audio, visual, and contextual data. As a result, they show limitations when dealing with complex fake news that relies on multimodal fusion, which impacts detection accuracy. To address this issue, an increasing number of studies have adopted multimodal fusion techniques, attempting to enhance detection by combining text, audio, and visual information. However, these methods still face challenges related to modal inconsistency, especially in the alignment of content between different modalities. For instance, some methods are able to combine text and comments for detection \cite{10.1145/3292500.3330935} or apply sentiment analysis \cite{ajao2019sentiment} to strengthen their judgments, but they fail to adequately consider the complex interaction between visual cues and emotions, resulting in room for improvement in detection performance. In addition, recent studies have incorporated more advanced techniques, such as counterfactual evidence and cross-modal reasoning, to improve model interpretability by introducing more diverse sources of evidence \cite{zhang2023rumor}. Although these methods have made progress in overcoming the limitations of multimodal models, they still face difficulties in handling dynamically manipulated videos, particularly with subtle visual and audio changes. Therefore, despite advancements in multimodal fusion and interpretability in video fake news detection, each method continues to face unique challenges: single-modal methods suffer from limited feature representation; multimodal methods struggle with modal inconsistency; and counterfactual and interpretability-based methods need further refinement to address complex manipulations in videos.

To address the challenges in fake news video detection, we propose VMID (Video Multimodal Information Detection), a multimodal framework that processes and integrates various data from short videos for input into a large language model (LLM). VMID utilizes pre-trained models—such as Whisper\cite{radford2022robustspeechrecognitionlargescale} for audio transcription, CogVLM2\cite{hong2024cogvlm2visuallanguagemodels} for visual frame analysis, and VSE (Video-subtitle-extractor)\cite{vse_github} for aligning textual and visual content—to create a unified multimodal representation. Furthermore, it incorporates metadata, including upload time, engagement metrics, and user comments, which provide valuable social context signals. The combined data is then structured into a prompt for evaluation by a LoRA (Low-Rank Adaptation)-tuned LLM, enabling the precise identification of misinformation within short videos.

Our contributions are threefold:

\begin{itemize} \item \textbf{Proposing an End-to-End Multimodal Fake News Detection Method (VMID)}: We propose a novel multimodal fake news detection method, VMID, which integrates visual, audio, and textual information from videos and processes them into a unified prompt for input into a LLM. VMID not only efficiently handles video content but also leverages social context signals, such as video metadata, to enhance the accuracy of fake news detection. Compared to existing methods, our approach comprehensively captures multimodal information from videos and classifies fake news with high precision.
\item \textbf{Full Implementation of the VMID System}: We provide a complete system implementation of the VMID method, adopting an end-to-end architecture that integrates data preprocessing, feature extraction, and inference through the LLM. This system can efficiently process large-scale data from short video platforms and perform automated fake news detection in real-world applications. With our implementation, the VMID system demonstrates scalability and ensures accurate analysis and reasoning across different modalities through its finely designed modules.
\item \textbf{Extensive Experimental Validation on the FakeSV Dataset}: We conducted extensive experiments on the FakeSV public dataset to evaluate the performance of VMID in fake news detection. The experimental results show that VMID achieves an improvement of approximately 9.87\% to 19.6\% in macro F1 score over other existing methods. We also compared VMID with advanced methods such as SV-FEND\cite{Qi2023FakeSV} and SVRPM\cite{wu2024interpretable}, and the results validate the superiority of our approach in multimodal fusion and social context utilization. These experiments further demonstrate the potential and effectiveness of VMID in large-scale real-world applications.
\end{itemize}
\section{Approach}
\label{sec:approach}
\begin{figure*}[h!]
    \centering
    \includegraphics[width=1\linewidth, trim=0 50 0 70, clip]{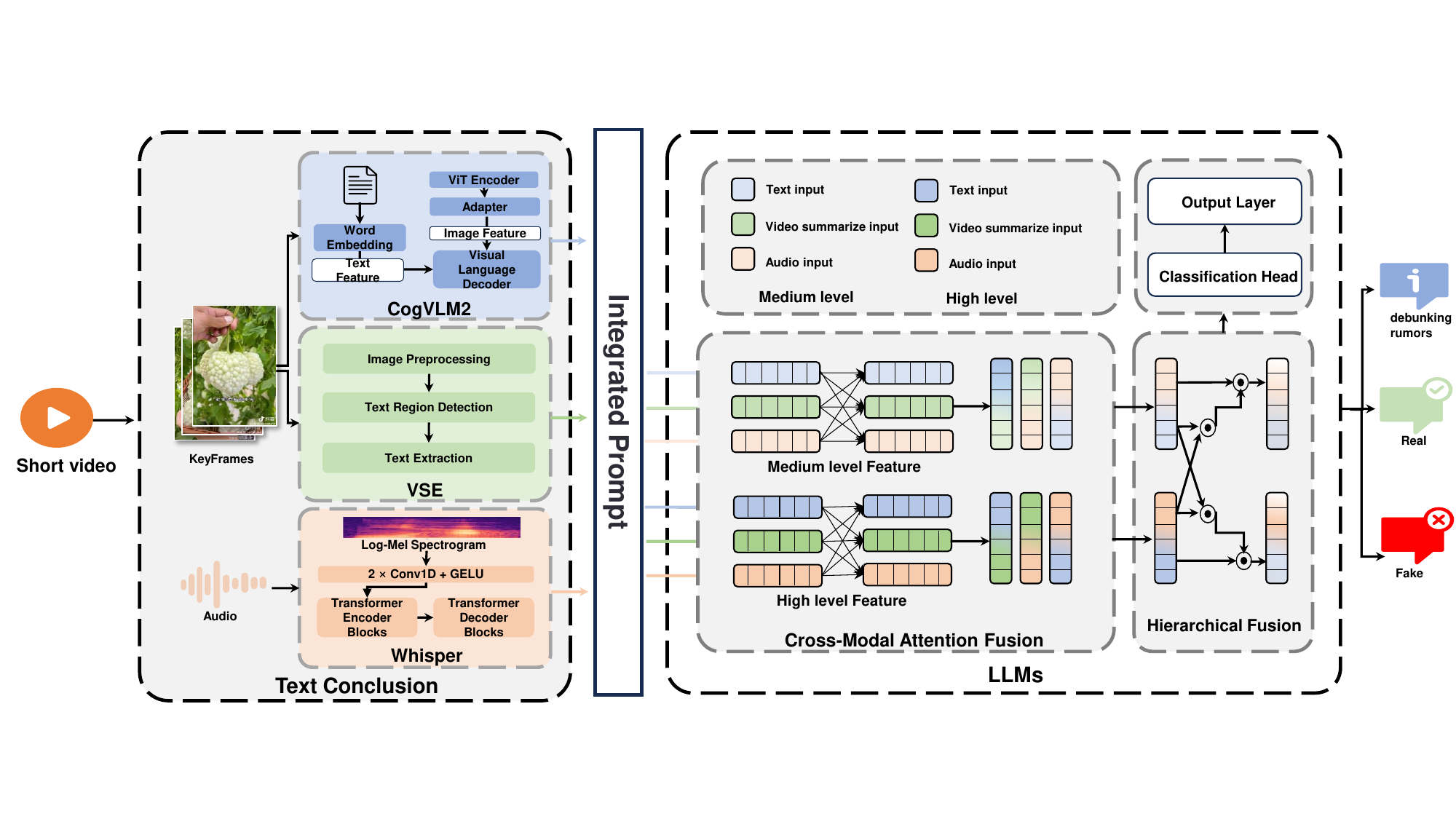}
    \caption{Architecture of the proposed framework VMID.The proposed approach integrates multiple modalities to generate text conclusions from short videos. It consists of three main components: CogVLM2, VSE, and Whisper. CogVLM2 processes keyframes extracted from the video, while VSE summarizes the video content. Whisper transcribes the audio component. These outputs are combined into an integrated prompt, which is fed into the LLMs. The LLMs employ cross-modal attention fusion at both medium and high levels to integrate information from text, video summaries, and audio inputs. Hierarchical fusion further combines these representations before passing them through a classification head and output layer to produce the final text conclusion.}
    \label{fig:Architecture}
\end{figure*}
In this section, we outline the overall framework for analyzing video content. As shown in Fig~\ref{fig:Architecture}, our framework abstracts the challenges of misinformation detection by integrating various data sources. We first introduce our problem formulation, which captures the complexities involved in identifying false information across different modalities. Next, we present the structure of our methodology, designed to effectively address these challenges. It is important to note that we describe our research approach at a high level, without delving into specific implementation details, as these may vary depending on the use case.

\subsection{Problem Formulation}

In this subsection, we formally define the core problem of misinformation detection by constructing a threat model and outlining the potential risks associated with the propagation of misinformation in short videos. The goal of this study is to utilize a multimodal framework to determine whether the content of short videos conveys misleading information or serves a debunking purpose.

Short videos have become a high-risk medium for misinformation due to their rapid dissemination and high user engagement. Misinformation can be spread through the manipulation of content, titles, and subtitles across various modalities, creating misleading narratives on both visual and linguistic levels. In light of this, the objective of this study is to design a multimodal framework capable of comprehensively assessing the role of short videos in the propagation or debunking of misinformation.

\paragraph{Data Collection}
To develop the misinformation detection model, multimodal data is extracted from video content, including subtitle text, transcribed audio, key visual elements from selected frames, and relevant metadata such as upload time and user interactions. Each data source offers distinct insights: subtitle text and transcribed audio provide linguistic content, key visual elements capture significant scenes within the video, and metadata (e.g., upload time and user interactions) offer contextual information regarding the video's publication and distribution. By integrating these multimodal data into a unified text input format, a comprehensive foundation is established for subsequent analysis.

\paragraph{Key Challenges}
In misinformation detection, we face the following main challenges, and we propose specific solutions for each:

\textbf{Challenge 1: Effective Multimodal Data Integration} — Integrating text, audio, visual, and metadata presents significant technical challenges due to variations in data format and relevance across modalities. To address this, we utilize the VMID framework to convert multimodal data into a unified text input format for efficient processing by the LLM.

\textbf{Challenge 2: Accurate Contextual Analysis} — Maintaining content intent consistency across different modalities to determine whether it propagates misinformation or serves as debunking content. The VMID framework integrates all modalities into a unified context to allow the LLM to accurately identify subtle contextual information.

\textbf{Challenge 3: Temporal and Spatial Consistency} — Ensuring that information extracted from different timestamps and visual frames remains contextually consistent. VMID integrates subtitle, audio transcript, and keyframe information into a single prompt to maintain temporal and spatial consistency across modalities.

 In conclusion, the VMID framework develops a comprehensive mechanism for video content analysis through the integration of multimodal data and the maintenance of contextual consistency, thereby improving the capability of the LLM to detect misinformation in short videos.

\subsection{Overall Structure of the Methodology}
The overall structure of the methodology, as shown in Fig~\ref{fig:Architecture}, consists of several key steps to enhance misinformation detection through multimodal data integration.

\subsubsection{Text Processing Workflow}

\begin{figure}[h!]
    \centering
    \includegraphics[width=\linewidth, trim=200 100 300 50, clip]{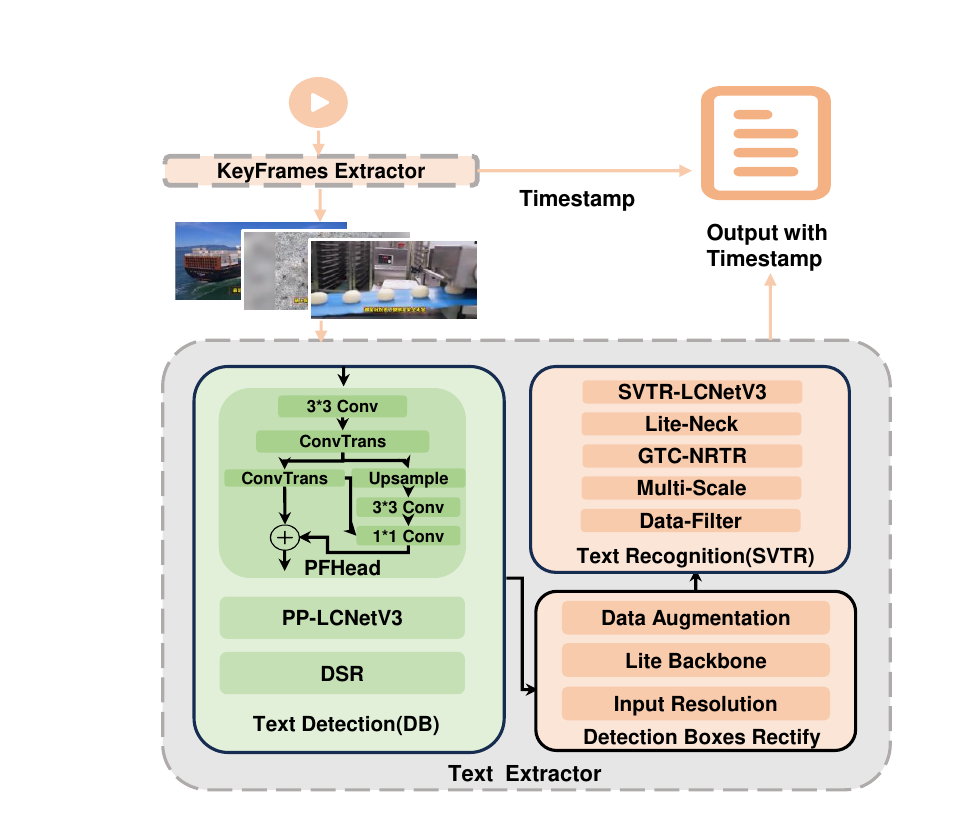}
    \caption{Overview of the Text Processing Workflow. The process begins with the KeyFrame Extractor, which captures critical moments from the video where subtitles appear. These frames are processed by the Text Detection module, featuring the PFHead structure for initial adjustments, the PP-LCNetV3 backbone for feature extraction, a Detection Boxes Rectify module to localize subtitle areas, and a Text Recognition (SVTR) module to extract text. The final output provides both the extracted subtitle text and its corresponding timestamps, ensuring temporal accuracy.}
    \label{fig:Subtitle_extracte}
\end{figure}

Figure~\ref{fig:Subtitle_extracte} illustrates the detailed workflow of the text processing pipeline, which includes keyframe extraction, text detection, feature extraction, subtitle area localization, text recognition, and output generation. Each step plays a vital role in ensuring that subtitles are accurately captured and recognized, thus enabling a comprehensive understanding of the video's content.

To maintain both accuracy and temporal context, our approach integrates timestamp-aligned subtitle extraction techniques. Further methods are presented in the language processing section to handle videos that do not contain embedded subtitles or rely on audio for content.

\paragraph{Subtitle Extraction}

The subtitle extraction process relies on Optical Character Recognition (OCR) techniques and consists of the following steps:

\begin{enumerate}
    \item \textbf{Keyframe Extraction}: The video is analyzed to extract keyframes that correspond to moments when subtitles appear. This step isolates the frames most relevant for subtitle detection and recognition, ensuring that only the most informative frames are processed.
    
    \item \textbf{Text Detection}: The extracted frames are passed through the Text Detection module, which utilizes the PFHead structure for preliminary adjustments. The PFHead module operates as follows: after the initial transposed convolution, the output is split into two branches. One branch undergoes upsampling and a 3x3 convolution to produce the final output, while the other branch progresses through an additional transposed convolution. The two branches are then merged via a 1x1 convolution, and the sum of the 1x1 convolution output and the second transposed convolution output yields a probability map that indicates potential text regions.
    
    \item \textbf{Feature Extraction}: The processed images are passed to the PP-LCNetV3 backbone for feature extraction. PP-LCNetV3, an advanced extension of the PP-LCNet family, combines high precision with efficient inference, making it suitable for various downstream tasks. Key optimizations include a learnable affine transformation module, enhanced re-parameterization techniques, an improved activation function, and modifications to network depth and width, all of which contribute to robust feature extraction for text detection and recognition.
    
    \item \textbf{Detection Box Rectification}: The features are passed to the Detection Box Rectification module, which refines the localization of text regions within each frame. To improve detection accuracy and robustness, we employ a Dynamic Scaling Ratio (DSR) strategy that progressively adjusts the scaling ratio during training. Specifically, as the training progresses, the scaling ratio \( s \) increases linearly from 0.4 to 0.6. The scaling ratio \( s \) at epoch \( t \) out of total epochs \( T \) is defined as:
    
    \begin{equation}
    s = 0.4 + \frac{0.2 \cdot t}{T}
    \label{eq:DSR}
    \end{equation}

    This strategy enhances the ability of the model to adapt to varying text sizes across different frames, improving localization accuracy and robustness.

    \item \textbf{Text Recognition}: The identified text areas are processed by the Text Recognition module, which uses a Transformer-based No-Recurrence Sequence-to-Sequence (NRTR) model. The recognition is trained using Connectionist Temporal Classification (CTC) loss, a popular approach for sequence-to-sequence tasks where alignment between input and output sequences is not guaranteed.

    The CTC loss function computes the probability \( p(Y|X) \) by summing over all possible alignments \( \pi \) between the input sequence \( X \) and the target sequence \( Y \), where \( \pi \) represents a valid alignment path. This can be written as:

    \begin{equation}
    \text{CTC Loss} = -\log \left( \sum_{\pi \in \text{Align}(X, Y)} p(\pi|X) \right)
    \label{eq:CTC}
    \end{equation}

    This formulation allows for flexible alignment, making it well-suited for OCR tasks where the text in video content can vary in length and position across frames. The advantages of using CTC include its ability to handle variable-length sequences, which is crucial for recognizing complex subtitle structures, and its improved recognition accuracy, particularly when combined with the NRTR model, which enhances performance for subtitles spanning multiple frames or with complex transitions.

    \item \textbf{Output Generation}: The system outputs the extracted text along with its corresponding timestamps, preserving both the textual content and its temporal alignment with the video.

\end{enumerate}

Through these steps, the subtitle extraction pipeline ensures accurate extraction and recognition of subtitles, contributing to a deeper understanding of the content within the video.

\subsubsection{Audio Processing}

We utilized Whisper\cite{radford2022robustspeechrecognitionlargescale} to process video audio, transcribing it into text to support natural language understanding tasks. This conversion from spoken to written form is essential for capturing semantic and emotional cues embedded in speech, such as tone, speech rate, and word choice, which convey the emotional state and intent of the speakers. By capturing these nuances, we enhance the depth and precision of our analysis.
\begin{itemize}
% \paragraph{Technical Details}

    \item \textbf{Audio Preprocessing}: We standardize the audio signal to a uniform format, typically a 16 kHz sampling rate, involving resampling and gain adjustment. Audio features are then extracted, including the Mel spectrogram, which is calculated as:

    \begin{equation}
    S(t, f) = \sum_k |X(k)|^2 \cdot h_m(t, f)
    \label{eq:audio}
    \end{equation}

    where \(S(t, f)\) is the Mel spectrogram value at time \(t\) and frequency \(f\), \(X(k)\) is the Fourier transform of the audio signal, and \(h_m(t, f)\) represents the Mel filter. This step improves recognition by enhancing critical frequency features in the audio.

    \item \textbf{Feature Encoding}: The extracted audio features are input to a Transformer-based encoder, which converts them into contextual embeddings by capturing long-range dependencies in the sequence. Using multi-head self-attention, the encoder maps the features through query (\(Q\)), key (\(K\)), and value (\(V\)) matrices, described by:

    \begin{equation}
    \text{Attention}(Q, K, V) = \text{softmax}\left(\frac{QK^T}{\sqrt{d_k}}\right)V
    \label{eq:Attention}
    \end{equation}

    where \(Q\), \(K\), and \(V\) are learned linear transformations of the audio feature embeddings, and \(d_k\) is the dimensionality of the key vectors. In this context, the feature sequence, initially representing the raw characteristics of the audio signal, is transformed through the \(Q\), \(K\), and \(V\) matrices, which capture the relationships between different segments of the audio sequence. Specifically, \(Q\) represents the elements that “query” others, \(K\) contains the “keys” providing potential matches, and \(V\) stores the “values” or responses. This process allows the model to focus on relevant temporal relationships, such as rhythm and pauses in speech.

    This mechanism enables the system to dynamically assign weights to different parts of the audio, prioritizing sections that convey meaningful information (e.g., intonation changes indicating emotional cues). In this way, the attention mechanism enhances the capability of the model to accurately interpret the context of each audio segment, leading to a deeper understanding of the spoken content.

    \item \textbf{Decoding}: A Transformer-based decoder predicts the most probable text sequence using beam search:

    \begin{equation}
    \text{BestSequence} = \arg\max_{\text{seq} \in \text{C}} \text{score}(\text{seq})
    \label{eq:BestSequence}
    \end{equation}

    where \(\text{C}\) represents possible decoding sequences, and \(\text{score}\) evaluates the likelihood of each sequence, ensuring optimal transcription accuracy.

    \item \textbf{Multi-task Training}: Whisper employs a unified prompt format to specify language and task with special markers, enhancing multilingual and multi-task adaptability by utilizing a large-scale supervised dataset.
\end{itemize}

Through these processes, Whisper surpasses traditional audio processing methods, delivering a unified representation of diverse audio modalities and establishing a reliable data foundation for in-depth analysis.

\subsubsection{Image Processing}

For the visual component of videos, particularly keyframes and video segments, we use the CogVLM2 model \cite{hong2024cogvlm2visuallanguagemodels}. It adeptly understands and describes visual content, creating detailed text descriptions that capture the essential information of the video. Carefully selected keyframes represent the most significant parts of the video, effectively transforming visual elements into meaningful textual descriptors. These keyframes highlight important scene changes, emotional expressions, or information transmission, providing a more intuitive understanding for viewers. The strength of CogVLM2 lies in its deep learning capabilities, which enable it to identify and parse complex visual features, generating contextually relevant descriptions.

\paragraph{Technical Details and Mathematical Principles}

\textbf{Keyframe Selection:}
In video processing, selecting representative keyframes is crucial. We analyze video frame rates and scene changes, considering factors such as scene transitions and emotional expressions, to select the most representative keyframes. Specifically, we use two methods:
1. \textbf{Frame Difference Method:} We calculate the pixel value differences between adjacent frames and select frames with significant differences as keyframes. If the pixel difference between two frames exceeds a predefined threshold, the frame is considered to contain important information changes.
2. \textbf{Motion Vector Analysis:} We use optical flow estimation techniques to analyze the direction and speed of object movements in the video. Frames with significant motion changes are selected as keyframes. This involves computing the motion vectors of objects in each frame and selecting frames with notable changes in these vectors.

\textbf{Visual Feature Extraction:}
We utilize the CogVLM2 model to extract complex visual features from keyframes, including objects, actions, and colors. The process involves:
1. Inputting keyframes into a pre-trained deep learning model, which extracts high-level features through multiple layers of convolutional neural networks (CNN).
2. Using a Vision Transformer (ViT\cite{dosovitskiy2021imageworth16x16words}) encoder to segment the image into fixed-size patches. Each patch is treated as an individual token, and positional embeddings are added to maintain positional information. These patch tokens interact through a multi-head self-attention mechanism to generate the final feature representation.

\textbf{Text Description Generation:}
Based on the extracted visual features, we generate detailed text descriptions to capture the core information of the video, providing viewers with a more intuitive understanding. The process includes:
1. Inputting the extracted visual features into a natural language generation model, which generates descriptive text using multiple layers of recurrent neural networks (RNN) or transformers.
2. Utilizing a cross-attention mechanism during the description generation process to integrate visual and linguistic features, helping the model focus on the most relevant parts of the image for accurate and rich descriptions.

\textbf{Multimodal Fusion:}
\begin{itemize}
    \item \textbf{Multimodal Adapter:} A 2x2 convolutional layer and the SwiGLU module align visual features with linguistic representations, achieving nearly lossless transformation. The SwiGLU module is defined as:
      \[
      \text{SwiGLU}(x) = x \odot \sigma(W_1 x + b_1) + W_2 x + b_2
      \]
      where \(W_1\) and \(W_2\) are weight matrices, \(b_1\) and \(b_2\) are bias terms, \(\sigma\) is an activation function (commonly ReLU), and \(\odot\) denotes element-wise multiplication. The SwiGLU module helps align visual features with linguistic representations, ensuring nearly lossless transformation. Specifically, it combines the input \(x\) with a gating activation function \(\sigma(W_1 x + b_1)\) to control the flow of information. The gating activation function determines which parts of the input should pass through, and the element-wise multiplication \(\odot\) ensures that only relevant features are retained. This mechanism allows for efficient and effective integration of visual and linguistic features, enhancing the model's ability to generate rich and accurate descriptions.
    \item \textbf{Multimodal Description Generation:} The multimodal adapter fuses visual features with linguistic features to generate comprehensive descriptions, integrating both visual information and linguistic context for richness.
\end{itemize}

\subsubsection{Metadata Extraction}

We gather additional metadata, such as video upload time, comment counts, like counts, and author information. This metadata plays a crucial role in contextualizing the content and understanding viewer engagement. For instance, the upload time can indicate the recency of information, while comment counts may reflect audience interest and perception. By analyzing this metadata, we gain deeper insights into how videos are perceived and the factors influencing viewer engagement.

\subsubsection{Multi-modal Integration}
After extracting information from each modality, all processed data is integrated into high-dimensional vectors and input into a LLM for comprehensive analysis. This integration allows the model to leverage its extensive knowledge base and reasoning capabilities to assess whether the video content propagates misinformation or serves to debunk false claims.

The integration of various modality vectors can be mathematically represented as follows:

\begin{equation}
\mathbf{V}_{\text{combined}} = \text{Concat}(\mathbf{V}_{\text{text}}, \mathbf{V}_{\text{audio}}, \mathbf{V}_{\text{visual}})
\label{eq:SwiGLU}
    \end{equation}

where \(\mathbf{V}_{\text{combined}}\) represents the combined high-dimensional vector, and \(\mathbf{V}_{\text{text}}\), \(\mathbf{V}_{\text{audio}}\), and \(\mathbf{V}_{\text{visual}}\) denote the vectors obtained from the text, audio, and visual modalities, respectively.

Once the vectors are combined, they undergo self-attention processing in the LLM, which can be expressed as Equation~\ref{eq:Attention}.

Finally, the output is generated through a decoder, which translates the model's hidden states into a natural language classification decision.

\subsubsection{Performance Optimization}

To further enhance model performance, we employ LoRA fine-tuning techniques and multi-modal feature fusion methods. These techniques have been extensively applied in our experiments, significantly improving the model's precision in identifying potential misinformation and enhancing its adaptability to complex content scenarios.

\paragraph{Technical Details and Mathematical Principles}

\textbf{LoRA Fine-Tuning:} LoRA is a technique for fine-tuning pre-trained models by incorporating low-rank matrices into the original model parameters. This method maintains the primary parameters of the model while updating only a small set of newly added parameters, thereby reducing computational costs and minimizing the risk of overfitting.

For each original weight matrix \( W \), LoRA adds two low-rank matrices \( A \) (column matrix) and \( B \) (row matrix). The updated weights can be expressed as:
\begin{equation}
W_{\text{new}} = W + \lambda \cdot A \cdot B
\label{eq:LoRA}
\end{equation}
where \( \lambda \) is a scalar that controls the intensity of the update. In our experiments, we first load the pre-trained model weights \( W \) and initialize the low-rank matrices \( A \) and \( B \). By adjusting \( \lambda \), we gradually update the model parameters during the fine-tuning process, reducing computational costs and avoiding overfitting. We initialize the low-rank matrices \( A \) and \( B \) with small random values and only update these matrices during training, keeping the original weights \( W \) unchanged. We use cross-validation to tune \( \lambda \) and employ cosine annealing schedulers and learning rate warm-up techniques to dynamically adjust the learning rate, ensuring the stability and efficiency of the training process.

\textbf{Multi-Modal Feature Fusion:} To comprehensively capture the information in video content, we adopt multi-modal feature fusion methods. Specifically, we use pre-trained models to extract descriptions from the video’s text, audio, and visual modalities. For instance, we use a pre-trained speech recognition model to convert audio into text and a pre-trained image recognition model to generate text descriptions of visual content. These descriptions are combined into a single fused text prompt, which is then input into a LLM for comprehensive analysis. The LLM processes this fused text prompt, converting it into a high-dimensional vector representation and performing multi-modal fusion within the model. To ensure consistency, a multi-modal adapter adjusts the spatial or channel dimensions of the features, allowing for seamless integration and analysis.

By training, we learn the weight coefficients \( \alpha \) and \( \beta \) to determine the importance of different modality features, ultimately generating a comprehensive feature vector \( f_{\text{fusion}} \).

Through the adoption of these methods and techniques, we not only enhance the performance of the model but also significantly improve its capability to identify potential misinformation through the integration of LoRA technology and multi-modal information. This ensures high accuracy and adaptability even in complex content scenarios.

\section{Implementation}
\label{sec:implementation}

This section details the implementation of our multimodal framework for short video content analysis. Figure \ref{fig:Architecture} illustrates the overall architecture, which integrates text, audio, and visual processing modules for comprehensive multimodal fusion and misinformation detection.

\subsection{Text Processing Enhancement}
We enhanced text processing by optimizing the VSE system to handle various subtitle formats and encodings, improving the accuracy of text extraction. In cases where subtitles are unavailable, Whisper transcription, combined with noise reduction preprocessing, significantly enhances transcription quality. To further analyze the extracted text, we apply advanced NLP (Natural Language Processing) techniques such as lemmatization, part-of-speech tagging, and named entity recognition. These techniques enrich the semantic analysis of the language and support subsequent sentiment and topic modeling.

\subsection{Audio Processing}
For audio processing, we utilized the Whisper pre-trained model, leveraging its robust multilingual transcription capabilities without requiring additional fine-tuning. This model enhances efficiency by reliably transcribing audio content directly, ensuring high accuracy across various languages and contexts.

\subsection{Visual Processing Enhancements}
Keyframe extraction is essential for accurately capturing video content. In CogVLM2, we employed a segment-based processing technique and a dynamic frame sampling strategy:

\subsubsection{Segment Processing Technique}
We divided each video into segments of a set duration $\text{segment\_duration\_seconds}$ seconds to ensure diverse key information is captured. The number of segments is calculated as follows:

\begin{equation}
\text{segments} = \lceil \frac{\text{total\_duration}}{\text{segment\_duration\_seconds}} \rceil
\label{eq:segments}
\end{equation}

where $\text{total\_duration}$ represents the total length of the video, and $\text{segment\_duration\_seconds}$ indicates the duration of the segment.

\subsubsection{Frame Selection Strategy}
Within each segment, we employed various frame selection strategies to capture keyframes:

\begin{itemize}
    \item \textbf{Uniform Frame Selection}: A base strategy that selects frames at uniform intervals. The formula for selecting frame IDs is:

    \begin{equation}
    \text{frame\_ids} = \text{np.linspace}(
    F_{\text{a}}, 
    F_{\text{b}} - 1, 
    N_{\text{f}}, 
    \text{dtype=int}
    )
    \label{eq:frame_selection}
    \end{equation}

    where \( F_{\text{a}} \) and \( F_{\text{b}} \) denote the starting and ending frames of the segment, and \( N_{\text{f}} \) is the number of frames to select.

    \item \textbf{Timestamp-based Matching}: We also used a timestamp-based approach, selecting frames that closely match timestamps to ensure no key information is missed.
\end{itemize}

\subsubsection{Similar Frame Filtering}
To reduce redundancy, we implemented a cosine similarity-based filtering mechanism. This method compares adjacent frames and retains only those with cosine similarity below a predefined threshold $\text{filter\_threshold}$. The similarity calculation is as follows:

\begin{equation}
\text{similarity} = \frac{\text{frame}_i \cdot \text{frame}_{i-1}}{\|\text{frame}_i\| \|\text{frame}_{i-1}\|} 
\label{eq:similarity}
\end{equation}

Frames with similarity above $\text{filter\_threshold}$ are discarded, ensuring that only unique frames contribute to visual analysis.

\subsection{Metadata Processing}
We developed a robust API to collect metadata, including video upload times, comment counts, likes, and user profile information. This metadata enables a deeper understanding of social context by analyzing user engagement patterns and behavior, which are essential for assessing credibility and identifying misinformation.

\subsection{Multimodal Fusion and Weight Optimization}
The final multimodal fusion integrates text, audio, and visual data into a unified format, fed into an LLM fine-tuned with LoRA for comprehensive analysis. To enhance fusion accuracy, we implemented a weight optimization strategy that dynamically allocates importance to each modality based on contextual relevance, allowing the model to prioritize the most informative modality in different scenarios. This adaptive weighting improves misinformation detection by enhancing the sensitivity of the model to context-specific signals.

By focusing on these enhancements, our framework demonstrates the practical effectiveness of multimodal misinformation detection in short video content, providing a reliable foundation for real-world applications.

\section{Performance Evaluation}
\label{sec:exp}
\begin{figure*}[!h]
    \centering
    \includegraphics[width=1\linewidth, trim=0 35 0 35, clip]{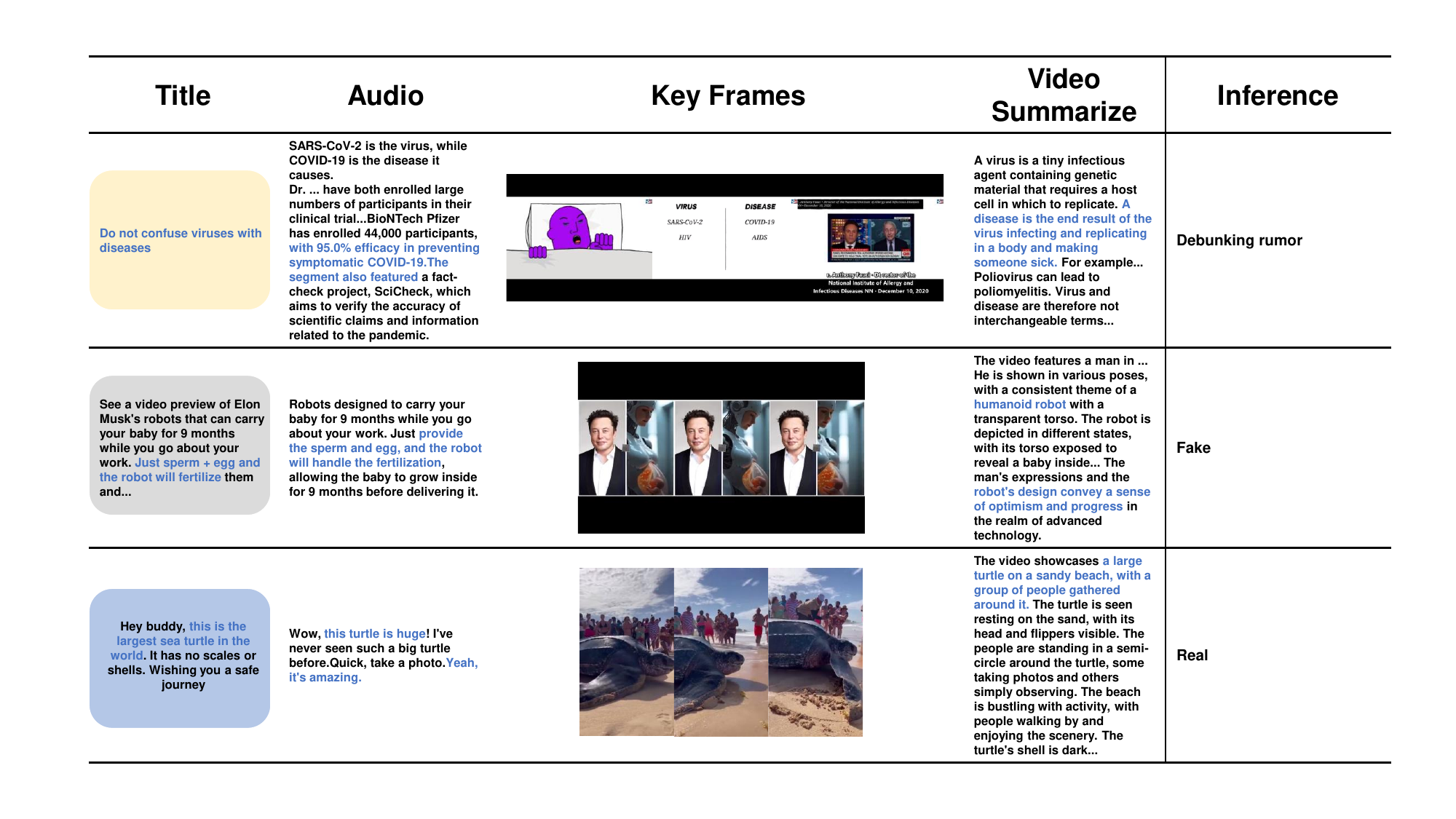}
    \caption{The short video detection examples presented are differentiated by background colors to represent the nature of the video content. Videos with a yellow background correspond to content aimed at debunking rumors, those with a gray background denote fake content, and videos with a blue background indicate genuine content. Additionally, the "Video Summary" column in the table presents the content following image processing. Relevant information that contributes to the inference process is highlighted in blue text.}
    \label{fig:cases-version2}
\end{figure*}
In this section, we present a comprehensive account of the experiments designed to evaluate the effectiveness of the proposed VMID method for detecting fake news videos. This includes an in - depth look at the baseline methods, the evaluation metrics employed, the experimental results obtained, and a detailed comparative analysis against existing models.

\subsection{Baseline Methods}
To thoroughly assess the performance of the VMID approach and establish a comprehensive benchmark, we compare it with a set of well-established baseline methods that utilize multiple modalities.
\subsubsection{MultiModality}
We use several existing state-of-the-art methods as multimodal baselines:
\begin{itemize}
\item \textbf{Hou et al 2019}\cite{10.1145/3340555.3353763}: Use the linguistic features from the speech text, acoustic emotion features, and user engagement features and a linear kernel SVM (Support Vector Machine) to distinguish the real and fake news videos. This method comprehensively considers information from multiple aspects for judging real and fake news.
\item \textbf{Medina et al 2020}\cite{medina-serrano-etal-2020-nlp}: Extract tf-idf vectors from the title and the first hundred comments and use traditional machine learning classifiers including logistic regression and SVM. It attempts to identify fake news by processing text information.
\item \textbf{Choi and Ko 2021}\cite{10.1145/3459637.3482212}: Use the topic distribution difference between title and comments to fuse them, and concat them with the visual features of keyframes. An adversarial neural network is used as an auxiliary task to extract topic-agnostic multimodal features for classification. This method focuses on the fusion and feature extraction of different modal information.
\item \textbf{Shang et al 2021}\cite{shang2021multimodal}: Use the extracted speech text to guide the learning of visual object features, use MFCC features to enhance the speech textual features, and then use a co-attention module to fuse the visual and speech information for classification. It emphasizes the interaction and information integration between different modalities.
\end{itemize}
In addition to the above, we also have the following notable multimodal baseline methods:
\begin{itemize}
\item \textbf{SV-FEND}\cite{Qi2023FakeSV}: This method capitalizes on multimodal information by integrating text, images, and user comments for fake news video detection. It has demonstrated robust performance and an outstanding ability to capture crucial multimodal cues, making it a significant benchmark in the field.
\item \textbf{SVRPM}\cite{wu2024interpretable}: SVRPM detects misleading content in short videos through modality tampering analysis. It also offers interpretability by highlighting discrepancies across different modalities, thereby facilitating effective misinformation identification.
\end{itemize}

\subsection{Evaluation Metrics}
Through this backtracking analysis, peak attention aligns with the item most pertinent to the query.

We adopted several key evaluation metrics to accurately measure the performance of each model under consideration:

\begin{itemize}
    \item \textbf{Accuracy (ACC)}: Computed as the ratio of correctly predicted samples to the total number of samples, this metric provides a fundamental measure of a model's performance.
    \item \textbf{Precision (Prec)}: Defined as the proportion of true positive predictions among all positive predictions made by the model, it offers insights into the model's ability to correctly identify positive instances without excessive false positives.
    \item \textbf{Recall (Recall)}: Representing the proportion of true positive predictions to the total number of actual positive cases, this metric reflects the capacity of the model to identify all relevant positive instances.
    \item \textbf{F1 Score}: The harmonic mean of precision and recall strikes a balance between the two, providing a comprehensive assessment of performance in handling both false positives and false negatives.
\end{itemize}

\subsection{Experimental Results}
The performance results of the different models in the fake news video detection task are presented in Table~\ref{tab:results01}. As evident from the results, our VMID method significantly outperforms all baseline models across the evaluated metrics.

% \begin{table}[htbp]
%     \centering
%     \resizebox{0.5\textwidth}{!}{
%     \begin{tabular}{ccccc}
%     \hline
%          Method                & Accuracy            & F1 Score       & Precision          & Recall \\ \hline
%          Hou et al., 2019\cite{10.1145/3340555.3353763}      & 71.89\%  & 71.29\%  & 73.88\%   & 71.89\%   \\ \hline
%         Medina et al., 2020\cite{medina-serrano-etal-2020-nlp}   & 75.58\%  & 75.50\%   & 75.92\%    & 75.58\%    \\ \hline
%         Choi and Ko, 2021\cite{10.1145/3459637.3482212}     & 78.32\%     & 78.31\%   & 78.37\%     & 78.32\%   \\ \hline
%         Shang et al., 2021\cite{shang2021multimodal}    & 74.45\%    & 74.39\%  & 74.67\%   & 74.45\%    \\ \hline
%         SV-FEND\cite{Qi2023FakeSV}               & 81.05\%    & 81.02\%    & 81.24\%        & 81.05\%    \\ \hline
%         SVRPM\cite{wu2024interpretable}        & 79.34\%      & 78.55 \%     & 79.75\%       & 78.16\%      \\ \hline
%         VMID_{\text{Qwen2.5}}(Ours)       & \textbf{90.93\%}        & \textbf{90.89\%}      & \textbf{90.88\%}       & \textbf{90.93\%}   \\ \hline
%     \end{tabular}
%     }
%     \caption{Results (\%) of different methods on our short video rumor dataset and the FakeSV dataset.}
%     \label{tab:results01}
% \end{table}
\begin{table}[htbp]
    \centering
    \resizebox{0.5\textwidth}{!}{
    \begin{tabular}{ccccc}
    \hline
         Method                & Accuracy            & F1 Score       & Precision          & Recall \\ \hline
         Hou et al., 2019\cite{10.1145/3340555.3353763}      & 71.89\%  & 71.29\%  & 73.88\%   & 71.89\%   \\ \hline
        Medina et al., 2020\cite{medina-serrano-etal-2020-nlp}   & 75.58\%  & 75.50\%   & 75.92\%    & 75.58\%    \\ \hline
        Choi and Ko, 2021\cite{10.1145/3459637.3482212}     & 78.32\%     & 78.31\%   & 78.37\%     & 78.32\%   \\ \hline
        Shang et al., 2021\cite{shang2021multimodal}    & 74.45\%    & 74.39\%  & 74.67\%   & 74.45\%    \\ \hline
        SV-FEND\cite{Qi2023FakeSV}               & 81.05\%    & 81.02\%    & 81.24\%        & 81.05\%    \\ \hline
        SVRPM\cite{wu2024interpretable}        & 79.34\%      & 78.55 \%     & 79.75\%       & 78.16\%      \\ \hline
        VMID$_{\text{Qwen2.5}}$ (Ours)       & \textbf{90.93\%}        & \textbf{90.89\%}      & \textbf{90.88\%}       & \textbf{90.93\%}   \\ \hline
    \end{tabular}
    }
    \caption{Results (\%) of different methods on our short video rumor dataset and the FakeSV dataset.}
    \label{tab:results01}
\end{table}

1) The \textbf{SV-FEND} model, with an accuracy of 81.05\%, presents a strong performance. However, VMID eclipses this with an impressive accuracy of \textbf{90.93\%} and an F1 score of \textbf{90.89\%}. This superiority showcases the enhanced capability of VMID in capturing and leveraging essential multimodal cues for accurate fake news detection, highlighting its advanced design and effectiveness.

2) The \textbf{FakeSV} model, despite its significance in highlighting the challenges associated with complex multimodal datasets through its average accuracy below 0.8, pales in comparison to VMID. The consistent accuracy of \textbf{90.93\%} and F1 score of \textbf{90.89\%} vividly demonstrate the robustness and superiority of VMID in handling the complexities inherent in such datasets.

3) Other baseline models, such as \textbf{Hou et al., 2019} and \textbf{Medina et al., 2020}, exhibit accuracies not exceeding 76\%, clearly falling short of VMID. These results spotlight the limitations of traditional feature extraction and classification techniques when faced with the intricate nature of complex multimodal data, further highlighting the innovative approach of VMID.

4) Notably, VMID not only achieves high accuracy but also excels in precision (\textbf{90.88\%}) and recall (\textbf{90.93\%}). This comprehensive performance underlines the prowess of VMID in the domain of fake news video detection. In contrast, the \textbf{SVRPM} model, with an accuracy of only \textbf{79.34\%}, reveals limitations in modality tampering detection, further highlighting the advantage of VMID.

\subsubsection{Loss Analysis}
\begin{figure}[!h]
    \centering
    \includegraphics[width=1\linewidth]{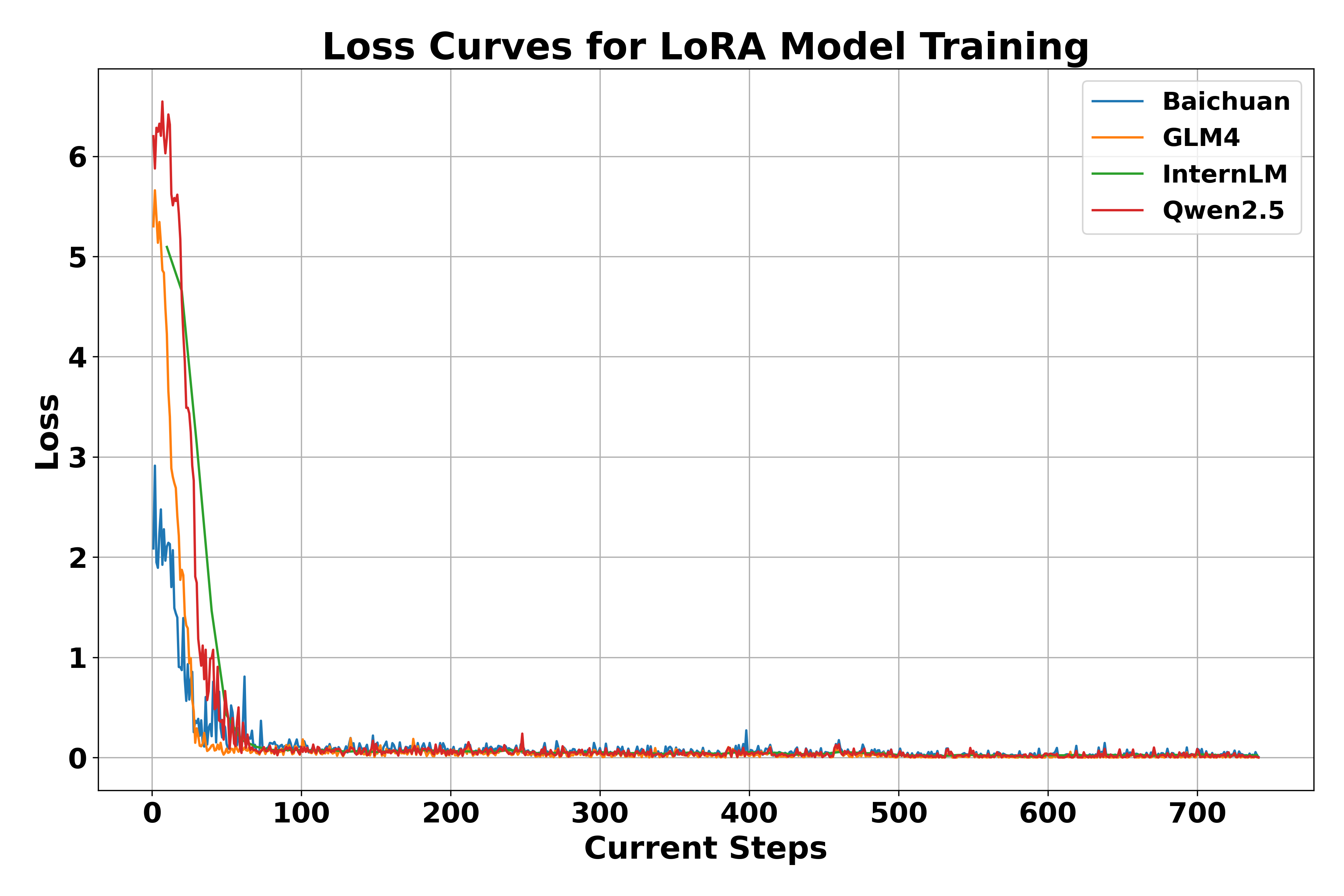}
    \caption{Loss curves for fine-tuning with LoRA on various models: Baichuan\cite{baichuan2}, GLM4\cite{glm4}, InternLM\cite{internlm}, and Qwen2.5\cite{qwen2.5}. The curves illustrate the training loss over time, demonstrating the convergence patterns of each model during the fine-tuning process.}
    \label{fig:loss_curve}
\end{figure}
\label{sec:loss_analysis}
To better understand the training dynamics and the effectiveness of the VMID method, we plot the loss curves during the training process. The loss curves provide insights into the convergence behavior of the model and highlight how well the model is learning over time. As shown in Figure~\ref{fig:loss_curve}, the loss steadily decreases over the training epochs, indicating that the model is progressively improving its performance. The sharp decline in the early stages of training suggests that the model is quickly learning the key features, while the more gradual reduction later on shows a fine-tuning phase where the model stabilizes. The plot demonstrates that VMID effectively optimizes its parameters for the task of fake news detection, further supporting the results reported in Fig\ref{fig:loss_curve}.

Moreover, we investigated the performance of different VMID models trained with various LLM architectures on the FakeSV dataset, as presented in Table~\ref{tab:results02}. The results demonstrate that VMID maintains a relatively high level of performance regardless of the specific LLM used.

\begin{table}[htbp]
    \centering
    \resizebox{0.5\textwidth}{!}{
    \begin{tabular}{ccccc}
    \hline
         LLM                & Accuracy            & F1 Score       & Precision          & Recall \\ \hline
         VMID$_{\text{Qwen2.5}}$   & \textbf{90.93}    & \textbf{90.89}   & \textbf{90.88}  & \textbf{90.93}            \\ \hline
         VMID$_{\text{GLM4}}$   &  \textbf{91.20} & \textbf{91.15} & \textbf{91.12} & \textbf{91.22}            \\ \hline
         VMID$_{\text{InternLM2.5}}$    & \textbf{88.48}    & \textbf{88.83}   & \textbf{88.92}  & \textbf{88.76}         \\ \hline
         VMID$_{\text{Baichuan}}$    & \textbf{88.30} & \textbf{88.35} & \textbf{88.38} & \textbf{88.32}            \\ \hline
    \end{tabular}
    }
    \caption{Performance comparison (in percentage) of different VMID models trained with various LLM architectures on the FakeSV dataset}
    \label{tab:results02}
\end{table}

Specifically, as shown in Table~\ref{tab:results02}, the VMID model trained with \textbf{GLM4} achieves the highest performance on the FakeSV dataset, with an accuracy of \textbf{91.20\%}, an F1 score of \textbf{91.15\%}, a precision of \textbf{91.12\%}, and a recall of \textbf{91.22\%}. Even when using other LLMs such as \textbf{InternLM2.5} and \textbf{Baichuan}, the VMID models still maintain high accuracies (88.48\% and 88.30\%, respectively), along with competitive F1 scores, precisions, and recalls. This indicates that the architecture of VMID is highly adaptable and robust, enabling effective utilization of different LLMs for fake news video detection on the FakeSV dataset. In conclusion, the VMID system demonstrates a clear and decisive advantage in handling the challenges presented by the \textbf{FakeSV} dataset, outperforming existing multimodal detection methods in terms of both accuracy and overall effectiveness. The results firmly establish the ability of VMID to integrate multimodal features seamlessly, thereby enhancing its robustness and accuracy in the crucial task of fake news detection.

We provide highlight successful predictions in Fig~\ref{fig:cases-version2}. In the figure, three short video examples are shown.These examples demonstrate the effectiveness of our method in distinguishing between different types of video content, including debunking rumors, fake content, and real incidents.

\subsection{Case Studies}

% In Fig\ref{fig:cases}, we demonstrate the accuracy of VMID in recognizing video content as real, fake, or debunking. The first three columns (title, audio, and keyframes) serve as the model's input data. VMID analyzes these multimodal inputs and accurately outputs the corresponding results. Different video categories are distinguished by background colors: yellow represents debunking content, gray represents fake content, and blue represents real content, further highlighting the effectiveness of VMID in recognizing various types of video content.
In Fig. \ref{fig:cases-version2}, we demonstrate the effectiveness of the VMID system in classifying video content into three categories: real, fake, and debunking. The system utilizes four key inputs—title, audio, keyframes, and video summary—which are highly correlated and provide comprehensive cues for the inference process. Title, audio, and keyframes serve as the multimodal features of the video, while the "Video Summary," obtained through image processing, offers additional contextual information. Together, these inputs enable the model to gain a more thorough understanding of the video content during the classification process. By integrating these related cues, VMID is able to accurately identify the content type of the video.

\begin{figure}[!h]
    \centering
    \includegraphics[width=1\linewidth, trim=30 50 30 30, clip]{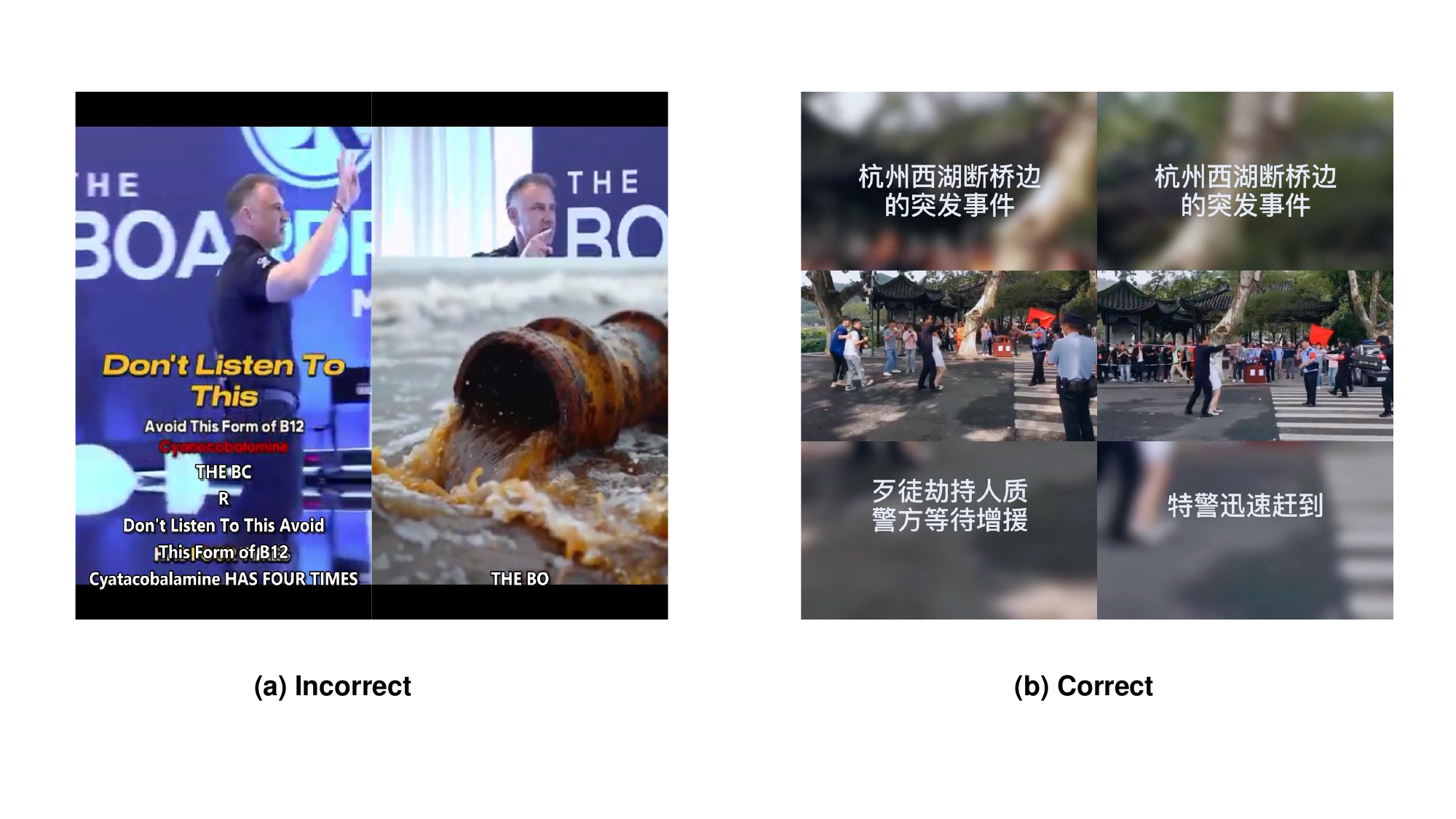}
    \caption{Two representative fake news videos from FakeSV, one detected and the other missed by VMID, showcasing its performance in fake news detection.}
    \label{fig:Cases Studies}
\end{figure}
% Fig \ref{fig:Cases Studies}(a) presents an example of a misclassified debunking video, titled "Indian park fitness equipment moves automatically at night; police investigate and suggest it may be caused by excessive lubricant." Despite the subtitle explaining the conclusion of the police, VMID incorrectly classifies this video as real content. Possible reasons include an insufficient number of debunking samples in the dataset, leading to reduced sensitivity to debunking content; limitations in multimodal fusion and contextual understanding, which hinder the ability to grasp complex causal relationships in the video; as well as noise in the input data, prior assumptions, and cultural differences that may introduce bias. This misclassification underscores the need for further improvements in multimodal fusion and contextual understanding.
Fig~\ref{fig:Cases Studies}(a) presents an example of a misclassified debunking video, titled "What Is It That American Regulation Agencies Actually Do?". The video claims that certain energy drinks contain cyanide from human sewage treatment plants and warns viewers to avoid supplements containing cyanocobalamin (a form of Vitamin B12). Despite the clear debunking nature of the content, VMID incorrectly classifies this video as real content. Possible reasons for this misclassification include an insufficient number of debunking samples in the dataset, which reduces the sensitivity of the model to debunking content; limitations in multimodal fusion and contextual understanding, which hinder the ability of the model to capture the exaggerations and sarcasm in the video; and noise in the input data or prior assumptions, which may lead to incorrect judgments. This misclassification highlights the need for further improvements in multimodal fusion and contextual understanding.

In contrast, Fig \ref{fig:Cases Studies}(b) showcases a successfully detected fake video, where a police drill was mistakenly described as a real event. By combining the knowledge base of the large language model with multimodal information, VMID effectively identified the misleading nature of the video, while SV-FEND\cite{Qi2023FakeSV}, without external knowledge support, failed to detect the error. This comparison demonstrates the advantage of VMID in addressing fake news detection in short videos by integrating multimodal information and LLM. These case studies not only demonstrate the effectiveness of VMID but also emphasize the importance of further optimizing multimodal analysis and contextual understanding in fake information detection.

\section{Related Work}

\subsection{Single-Modal Fake News Detection}

Early research on fake news detection primarily focused on single-modal information processing, such as text, images, or videos. These studies attempted to identify misinformation by analyzing user behavior and information dissemination patterns on social media platforms. For instance, (Papadopoulou et al., 2019)\cite{Papadopoulou2019ACO} proposed a feature-based SVM classifier that utilizes video metadata, linguistic features of titles, and the credibility of comments to identify rumors. Similarly, (Li et al., 2022)\cite{li2022cnn} and (Ran et al., 2022)\cite{ran2022mgat} constructed heterogeneous graph models incorporating social media user information, which demonstrated improved performance in rumor detection tasks. Furthermore, (Medina et al. 2020)\cite{serrano2020nlp} used tf-idf vectors based on titles and comments, combined with co-conspiratorial relationships among comments, to enhance detection accuracy. While these single-modal approaches laid the foundation for fake news detection, they exhibit limitations in integrating multiple cues to improve applicability and comprehensiveness. In particular, in the context of short videos, single-modal methods often fail to cover all relevant information sources, making the comprehensive analysis of multimodal data crucial for improving detection accuracy and robustness. In comparison, the VMID model integrates text, video, and social context information, thereby improving accuracy and performance in fake news detection.

\subsection{Multimodal Fake News Detection}

With the increasing prevalence of short video content, multimodal fake news detection has become a prominent research focus. Multimodal approaches can combine text, image, and video information, capturing richer data features to improve detection accuracy. In the domain of text-image fake news detection, several neural network architectures have been proposed to identify feature associations between text and images. For example, (Qi et al., 2021) \cite{10.1145/3474085.3481548} introduced a model that effectively integrates text and image information to enhance the processing of cross-modal data. (Choi and Ko, 2021)\cite{10.1145/3459637.3482212} combined topic distribution analysis with adversarial training to strengthen feature learning in the complex task of handling cross-modal data.

In the context of short video fake news detection, multimodal methods have made further advancements. For instance, the SV-Defend model proposed by (Qi et al., 2023)\cite{Qi2023FakeSV} significantly improved detection performance by extracting high-information cross-modal features and integrating social context. (Wu et al., 2024)\cite{wu2024interpretable} introduced a modality-tampering-based rumor detection approach, which identifies inconsistencies between video content and accompanying information through modality tampering detection and cross-modal matching pretraining tasks. Additionally, the approach employs interpretability mechanisms to trace the decision-making process, thereby enhancing the rationality of the detection results. Furthermore, (Zhang et al., 2021)\cite{ZHANG2021100005} proposed a framework for detecting super-spreaders based on WLAN (Wireless Local Area Network) logs, providing insight into how user behavior and social context can enhance detection. This approach, although focused on pandemic-related misinformation, is relevant to understanding the role of social context in fake news propagation. (Ranjan and Kumar, 2022)\cite{RANJAN2022100034} also highlighted the importance of user behavior analysis in distinguishing malicious users from legitimate ones, which could be integrated into multimodal detection systems. Despite these advancements, challenges remain in effectively fusing video, text, and social context information, particularly in handling the interdependencies and interactions between modalities to ensure comprehensive and accurate detection. The VMID model addresses these challenges by incorporating fine-grained cross-modal feature integration and information coordination, which enhances both the diversity and robustness of fake news detection.

\subsection{Applications of Large Language Models in Fake News Detection}

LLMs have demonstrated powerful interpretative capabilities, opening new possibilities for fake news detection. Since fake news is primarily disseminated via text on social media, LLMs such as GPT-2 (Zellers et al., 2019) \cite{NEURIPS2019_3e9f0fc9} and BERT (Singhal et al., 2020) \cite{singhal2020spotfake+} have been widely employed for both fake news generation and detection. Recently, ChatGPT has exhibited exceptional performance in understanding complex texts, showing substantial potential in the field of fake news detection. For example, (Huang et al., 2023) \cite{huang2024fakegptfakenewsgeneration} introduced the FakeGPT framework, by using rationale-aware prompting strategies to significantly improve the detection performance of ChatGPT, this approach highlights the potential of LLMs in fake news detection

Despite the strong performance of LLMs in text analysis, they face limitations when dealing with multimodal data such as video and images. In contrast, VMID overcomes these limitations by leveraging the strengths of LLMs in text processing while effectively integrating video and social context information, enabling the handling of more complex multimodal data. Furthermore, by utilizing deep integration of social context and video content, VMID addresses the bottlenecks of traditional LLMs in multimodal processing, further enhancing performance in fake news detection.

\section{Conclusion}

This paper introduces VMID, a novel multimodal model designed to tackle the challenges of fake news detection in short videos. By integrating various modalities—such as video content, metadata, and social context—VMID significantly improves the accuracy of fake news detection, overcoming key limitations of existing methods that primarily rely on single-modal approaches or inadequate fusion techniques. Extensive experiments on the FakeSV dataset show that VMID outperforms current models, achieving a significant improvement in the macro F1 score, highlighting its effectiveness and robustness.
As the problem of fake news continues to evolve across different domains, the need for reliable detection methods has become increasingly urgent. The success of VMID highlights the potential of LLMs in processing multimodal data, marking a significant advancement in the development of automated misinformation detection systems. Looking to the future, a promising research direction involves exploring LLMs for real-time, internet-based fact-checking. By utilizing live data queries from trusted sources or databases, these models could assess the authenticity of news content and provide dynamic, real-time evaluations. Such innovations are expected to further enhance detection accuracy and efficiency, enabling timely interventions to curb the spread of misinformation.
Furthermore, continuous improvements in model interpretability and the integration of additional data sources, such as user behavior and cross-platform metadata, will further strengthen the effectiveness of fake news detection systems. As the field of misinformation detection evolves, the combination of LLMs with real-time information retrieval has the potential to revolutionize the domain, offering more reliable and scalable solutions to address the growing challenges of fake news in the digital era.

\section*{Acknowledgment}
This study was partially supported by the National Natural Science Foundation of China (No. 62302266, 62232010, U23A20302), the Shandong Science Fund for Excellent Young Scholars (No.2023HWYQ-008), the project ZR2022ZD02 supported by Shandong Provincial Natural Science Foundation, the Humanities and Social Sciences Research Project of the Ministry of Education [24YJCZH459] and the Guangdong Province Philosophy and Social Science Planning Project [GD23XGL017].

\bibliographystyle{IEEEtran}
\bibliography{paper}

\end{document}